\def\year{2021}\relax
\documentclass[letterpaper]{article} % DO NOT CHANGE THIS
\usepackage{aaai21}  % DO NOT CHANGE THIS
\usepackage{times}  % DO NOT CHANGE THIS
\usepackage{helvet} % DO NOT CHANGE THIS
\usepackage{courier}  % DO NOT CHANGE THIS
\usepackage[hyphens]{url}  % DO NOT CHANGE THIS
\usepackage{graphicx} % DO NOT CHANGE THIS
\usepackage{natbib}  % DO NOT CHANGE THIS AND DO NOT ADD ANY OPTIONS TO IT
\usepackage{caption} % DO NOT CHANGE THIS AND DO NOT ADD ANY OPTIONS TO IT
\usepackage{calc,amsfonts,amssymb,amsmath,bm,color,theorem, dsfont}%, amsthm}
\usepackage{multirow}
\usepackage{subfig}
\usepackage{comment}
\usepackage{xspace}
\usepackage{wrapfig,lipsum,booktabs}
\usepackage{algorithmic}
\usepackage{epstopdf}
\usepackage{cite}

\DeclareMathAlphabet\mathbfcal{OMS}{cmsy}{b}{n}

\newtheorem{theorem}{Theorem}

\newtheorem{definition}{Definition}

\newcommand{\method}{{e}\textsc{Tree}\xspace}

\newcommand{\R}{\mathbb{R}}

\def\BibTeX{{\rm B\kern-.05em{\sc i\kern-.025em b}\kern-.08em
    T\kern-.1667em\lower.7ex\hbox{E}\kern-.125emX}}

\usepackage{xcolor}
\usepackage[linesnumbered,ruled]{algorithm2e}

\SetCommentSty{mycommfont}

\SetKwInput{KwInput}{Input}                % Set the Input
\SetKwInput{KwOutput}{Output}

\title{\method: Joint Nonnegative Matrix Factorization and Tree Structure Learning}
\title{\method: Learning Tree-structured Embeddings}
\author {
        Faisal M. Almutairi\textsuperscript{\rm 1,2},
        Yunlong Wang\textsuperscript{\rm 2},
        Dong Wang\textsuperscript{\rm 2},
        Emily Zhao\textsuperscript{\rm 2},
        and Nicholas D. Sidiropoulos\textsuperscript{\rm 3}\\
}
\affiliations{
    \textsuperscript{\rm 1}Department of ECE, University of Minnesota, MN, USA \\
    \textsuperscript{\rm 2}Advanced Analytics, IQVIA Inc.,  Plymouth Meeting, PA, USA\\
    \textsuperscript{\rm 3}Department of ECE, University of Virginia, VA, USA\\
    almut012@umn.edu, $\{$yunlong.wang, dong.wang, emily.zhao$\}$@iqvia.com,
    nikos@virginia.edu
}

\begin{document}
%\linenumbers  %

%\begin{comment}

\maketitle

\begin{abstract}
Matrix factorization (MF) plays an important role in a wide range of machine learning and data mining models. MF is commonly used to obtain item embeddings and feature representations due to its ability to capture correlations and higher-order statistical dependencies across dimensions. In many applications, the categories of items exhibit a hierarchical tree structure. For instance, human diseases can be divided into coarse categories, e.g., bacterial, and viral. These categories can be further divided into finer categories, e.g., viral infections can be respiratory, gastrointestinal, and exanthematous viral diseases. In e-commerce, products, movies, books, etc., are grouped into hierarchical categories, e.g., clothing items are divided by gender, then by type (formal, casual, etc.). While the tree structure and the categories of the different items may be known in some applications, they have to be learned together with the embeddings in many others. In this work, we propose \method, a model that incorporates the (usually ignored) tree structure to enhance the quality of the embeddings. We leverage the special uniqueness properties of Nonnegative MF (NMF) to prove identifiability of \method. The proposed model not only exploits the tree structure prior, but also learns the hierarchical clustering in an unsupervised data-driven fashion. We derive an efficient algorithmic solution and a scalable implementation of \method that exploits parallel computing, computation caching, and warm start strategies. We showcase the effectiveness of \method on real data from various application domains: healthcare, recommender systems, and education. 
We also demonstrate the meaningfulness of the tree obtained from \method by means of domain experts interpretation.        
\end{abstract}

%\vspace{-4mm}
\section{Introduction}
Matrix Factorization (MF) plays an important role in a wide range of machine learning models, for various applications such as dimensional reduction and embedding. 
A popular task is matrix completion, where the goal is to infer the unknown/missing matrix entries from the observed ones. A common approach is to employ MF to find a reduced-dimension representation (embedding) of each element corresponding to the matrix dimensions (e.g., users and items) These embeddings capture the essential information due to the ability of MF to capture  correlations and higher-order statistical dependencies across dimensions.  
The entry corresponding to the $i^{th}$ user and $j^{th}$ item can be inferred by the inner product of their embeddings. 
Matrix completion finds a wide range of applications including collaborative filtering in recommender systems \cite{koren2008factorization}, disease and treatment prediction and patient subtyping \cite{wang2019enhancing} in healthcare analytics, student performance prediction and course recommendation in learning analytics \cite{almutairi2017context}, and image processing \cite{liu2015low}.
\begin{figure}[t]
	\centering
	{\includegraphics[width=0.325\textwidth,height=2.2cm]{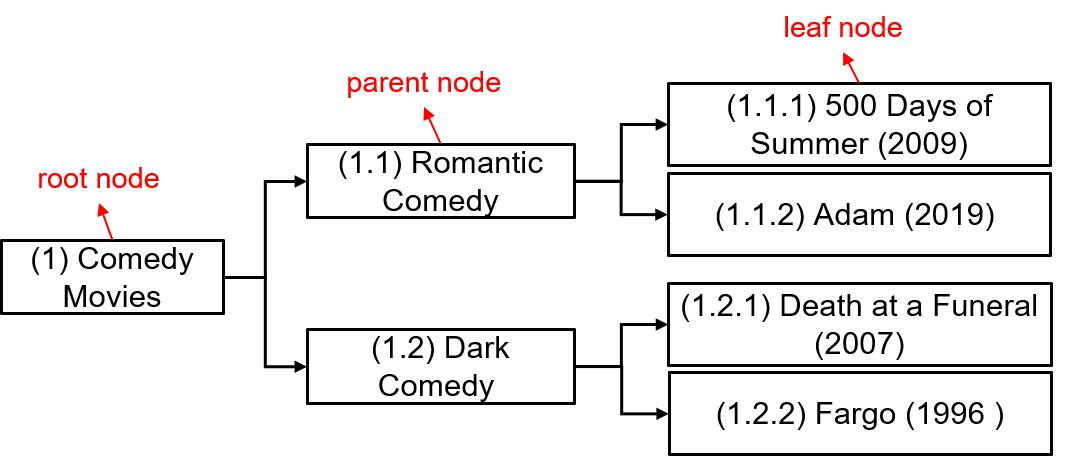}}
	\setlength{\belowcaptionskip}{-17pt}
	\vspace{-2mm}
	\caption{An example of hierarchical movie categories.}
	\label{fig:treeExp}
\end{figure}

Incorporating side contextual information or priors, e.g., sparsity \cite{hoyer2004non}, smoothness, and latent clustering \cite{yang2016learning}, is well-motivated in matrix factorization and completion of sparse data. This is because a major challenge stems from the fact that we aim to find latent representations from very few samples. 
In this paper, we present a principled approach that incorporates the \emph{unknown} implicit tree structure prior. 
In many applications, categories of items display a hierarchical tree structure. 
In higher education, for instance, courses form multiple trees via their prerequisite hierarchy. 
Movie genres, e.g., comedy, action, and fantasy, comprise different fine subcategories as illustrated in the example in Fig. \ref{fig:treeExp}.    
Another example appears in Electronics Health Records (EHR) in healthcare analytics, where medical service (diagnoses, procedures, and prescription) can be clustered into subcategories, and these subcategories can also be grouped into coarse categories (examples are provided in the experimental results in Fig. \ref{fig:Med_MCI}).   
Individuals, e.g., users, students, and patients, also exhibit hierarchical clusters where the common traits between people increase as we move down from the root nodes to the leaf nodes in a tree \cite{maleszka2013method, wang2014hgmf}.  
In many applications, the categorical hierarchy is either unknown, or requires manual labeling of massive amounts of data.

Incorporating tree structures in machine learning models has been recently considered, mostly in recommender systems \cite{nikolakopoulos2015hierarchical, li2019hhmf, zhong2012contextual} and also in other applications such as image processing \cite{fan2018matrix}, clustering and classification \cite{trigeorgis2016deep}, 
and natural language processing (NLP) \cite{shen2018ordered}.
For example, the recommender system model in \cite{yang2016learningHier} penalizes MF with the distance between users who share common traits {based on hierarchically-organized features.}
{In another MF model \cite{sun2017exploiting}, the item embeddings are assumed to form a tree, where each 
leaf node represents a single item and the
parent nodes contain subsets of items (categories). 
The embeddings of parent nodes and leaf nodes are learned jointly. The final item feature vector is modeled as a weighted sum of its embedding and those of the categories it belongs to.}
%The work in \cite{nikolakopoulos2015hierarchical} models a user preference as a probability distribution over a tree-structured item-space.
Regularizing MF with a pre-defined tree prior has been also explored in response prediction in online advertising \cite{menon2011response}.  
In this example, the tree groups the set of ads according to their campaigns, and the campaigns are further grouped based on the advertisers running them.

\emph{All} the aforementioned methods assume that the tree structure is known apriori, or learned separately via side information. 
Recently, \cite{wang2018exploring, wang2015exploring} proposed to capture the unknown implicit tree structure via a model based on nonnegative matrix factorization (NMF). 
In a three-layer tree, the embedding of a leaf node (item/user) is assumed to be a linear combination of \emph{all} the parent nodes (subcategories) in the intermediate layer, and each subcategory is a linear combination of all the categories in the root nodes.     
The weights that determine the memberships of a child node to the parent nodes are non-negative and learned by the model. 
This results in a fully connected tree, thus, a clear tree clustering can not be obtained. Moreover, \cite{wang2018exploring,wang2015exploring} imposes the implicit tree as a hard constraint, which can be restrictive if the data do not exactly follow the imposed prior.         

In this paper, we propose \method ({L\underline{e}arning \underline{Tree}-structured Embeddings}), a framework that integrates the unknown implicit tree structure into a low-rank nonnegative factorization model to improve the quality of embeddings.
\method does not require any extra information and jointly learns: i) the embeddings of all the tree nodes (items, subcategories, and main categories), and ii) the tree clustering in an unsupervised fashion.
Unlike \cite{wang2015exploring, wang2018exploring}, the obtained tree provides clear hierarchical clusters as each node belongs to exactly one parent node, e.g., an item belongs to one subcategory, and a subcategory belongs to one main category.  
The formulation of \method handles partially observed data matrices, which appear often in real-world applications. 
We derive an efficient algorithm to compute \method with a scalable implementation that leverages parallel computing, computation caching, and warm-start strategies.   
Our contributions can be summarized as follows: 

    \noindent {\bf 1. Formulation:} \method provides an intuitive formulation that: i) exploits the tree structure, and ii) learns the hierarchical clustering in an unsupervised data-driven fashion.   

\noindent {\bf 2. Identifiability:} We leverage the special uniqueness properties of NMF to prove identifiability of \method.

\noindent {\bf 3. Effectiveness:} \method significantly improves the quality of the embeddings in terms of matrix completion error on data from recommender systems, healthcare, and education.  
 
\noindent {\bf 4. Interpretability:} We demonstrate the meaningfulness of the tree clusters learned by \method using real-data interpreted by domain experts.

\vspace{-2mm}
\section{Background}
In this section, we provide the background needed before presenting  \method. We review NMF and related recent identifiability results, followed by a brief background on the alternating direction method of multipliers (ADMM).   

\noindent \underline{Notation:} $x$. ${\bf x}$, ${\bf X}$ denote scalars, vectors, and matrices, respectively; ${\bf X}(i,:)$ (${\bf X}(:,j)$) refers to the $i^{th}$ row ($j^{th}$ column) of ${\bf X}$; ${\bf X}(\mathcal{J},:)$ denotes the rows of ${\bf X}$ in the set $\mathcal{J}$. The product $\odot$ is the Hadamard (element-wise) product.          

\vspace{-1mm}
\subsection{Non-negative Matrix Factorization}
Assume we have a healthcare data matrix ${\bf X}\in \R^{N \times M}$ indexed by (patient, medical service), where ${\bf X}(i,j)$ denotes the number of times the $i^{th}$ patient has received the $j^{th}$ service. In other applications, ${\bf X}$ may contain the ratings given by users to items, or the grades received by students in their courses. In some parts of this paper, we refer to patients, users, students, etc. as individuals, and to medical services, products, etc. as items.
NMF models aim to decompose the data matrix into low-rank latent factor matrices as ${\bf X} = {\bf A}{\bf B}^T$, where ${\bf A}\in \R^{N \times R}$, ${\bf B}^{M \times R}$ only have non-negative values, and $R\leq\text{min}(N, M)$ is the matrix rank. NMF has gained considerably special attention as it tends to produce interpretable representations. For instance, it has been shown that the columns of ${\bf A}$ produce clear parts of human faces (e.g., nose, ears, and eyes) when NMF in applied on a matrix ${\bf X}$ whose columns are vectorized face images \cite{lee1999learning}. 
In practice, NMF is often formulated as a bilinear optimization problem:     
\begin{equation}\label{eq:NMF}
\begin{aligned}
\min_{{\bf A}\geq {\bf 0},{\bf B}\geq {\bf 0}} \quad & 
\frac{1}{2}
\mathcal{F}({\bf A}, {\bf B}) = \|{\bf W}\odot({\bf X} -  {\bf AB}^T)\|_F^2
\end{aligned}
\end{equation}
where ${\bf W} \in \{0,1\}^{N \times M}$ has ones at the indices of the observed entries in ${\bf X}$, and zeros otherwise. Each row of ${\bf A}$ corresponds to the embedding/latent representation of the corresponding individual, whereas the rows of ${\bf B}$ are the embeddings of the items.

\noindent{\bf Identifiability of NMF:}
The interpretability of NMF is intimately related to its 
uniqueness properties -- the latent factors are identifiable under some conditions (up to trivial ambiguity, e.g., scaling/counter-scaling or permutation).  
To facilitate our discussion of the uniqueness of \method, we present the following definitions and established identifiability results.  

\vspace{-3mm}
\begin{definition}[Identifiability]
The NMF of ${\bf X} = {\bf A}{\bf B}^T$ is said to be (essentially) unique if ${\bf X} = \widetilde{\bf A}\widetilde{\bf B}^T$ implies $\widetilde{\bf A} = {\bf A}{\bf \Pi}{\bf D}$ and $\widetilde{\bf B} = {\bf B}({\bf \Pi}{\bf D})^{-1}$, where ${\bf \Pi}$ is a permutation matrix, and ${\bf D}$ is a diagonal positive matrix. 
\end{definition}
\vspace{-5mm}
\begin{definition}[Sufficiently Scattered]\label{def:ss}
A nonnegative matrix ${\bf B} \in \R^{M \times R}$ is said to be sufficiently scattered if: 
1) cone$\{{\bf B}^T\}$ $\supseteq \mathcal{C}$, and 
2) cone$\{{\bf B}^T\}$ $\cap$ \text{bd}$\{\mathcal{C}^\star\} = \{\lambda{\bf e}_k|\lambda\geq0, k=1,\dots, R\}$, where $\mathcal{C} = \{{\bf x}| {\bf x}^T{\bf 1} \geq \sqrt{R-1}\|{{\bf x}}\|_2\}$, $\mathcal{C}^{\star} = \{{\bf x}| {\bf x}^T{\bf 1} \geq \|{{\bf x}}\|_2\}$, cone$\{{\bf B}^T\} = \{{\bf x}|{\bf x} = {\bf B}^T{\bm \theta},~\forall {\bm \theta} \geq {\bf 0}, ~{\bf 1}^T{\bm \theta} = 1\}$, 
and cone$\{{\bf B}^T\}^{\star} = \{{\bf y}|{\bf x} = {\bf x}^T{\bf y},~\forall {\bf x} \in cone\{{\bf H}^T\}\}$ are the conic hull of ${\bf B}^T$ and its dual cone, respectively, and bd is the boundary of a set.      
\end{definition}
\vspace{-2mm}

%\noindent 
The works in \cite{fu2015blind,lin2015identifiability}
prove that the so-called volume minimization (VolMin) criterion can identify the factor matrices 
if ${\bf A}$ is full-column rank, and the rows of ${\bf B}$ are sufficiently scattered (Definition \ref{def:ss}) and sum-to-one (row stochastic). 
Recently, Fu \emph{et al.} shifted the row stochastic condition on rows of ${\bf B}$ to the \emph{columns} of ${\bf B}$. 

\vspace{-3mm}
\begin{theorem}[NMF Identifiability] \cite{fu2018identifiability}\label{theo:NMF}
${\bf A}$ and ${\bf B}$ are essentially unique under the criterion of minimizing det$({\bf A}^T{\bf A})$ w.r.t. ${\bf A}\in \R^{N \times R}$ and ${\bf B} \in \R^{M \times R}$, subject to ${\bf X} = {\bf A}{\bf B}^T$ and ${\bf B}^T{\bf 1} = {\bf 1}, {\bf B} \geq {\bf 0}$
if ${\bf B}$ is sufficiently scattered, and rank(${\bf X}$) = rank(${\bf A}$) = $R$.
\end{theorem}
\vspace{-2mm}

%\noindent 
Theorem \ref{theo:NMF} provides an intriguing 
generalization of NMF, as it pertains to a more general factorization. Note that ${\bf A}$ is {\em not} restricted to be non-negative. Also note that the column-sum-to-one constraint on ${\bf B}$ is without loss of generality, as one can always
assume the columns of ${\bf B}$ are scaled by a diagonal matrix ${\bf D}$, and compensate for this scaling in the columns of ${\bf A}$, i.e., ${\bf X} = ({\bf A}{\bf D}^{-1})({\bf B}{\bf D})^T$.       

\vspace{-1mm}
\subsection{Alternating Direction Method of Multipliers}
ADMM is a primal-dual algorithm that solves convex optimization problems in the form
\vspace{-2mm}
\begin{equation}\label{eq:ADMM}
\begin{aligned}
\text{min}_{{\bf x},{\bf z}} \quad & 
f({\bf x}) + g({\bf z})\\
 \text{s.t.}  \quad &  {\bf A}{\bf x} + {\bf B}{\bf z} = {\bf c} 
\end{aligned}
\end{equation}
\vspace{-2mm}

\noindent by iterating the following updates
\vspace{-2mm}
\begin{equation}\label{eq:ADMM_updates}
\begin{aligned}
{\bf x} &\leftarrow \text{arg} ~\text{min}_{\bf x}\quad f({\bf x})+ \rho/2 \|{\bf A}{\bf x} + {\bf B}{\bf z} - {\bf c} + {\bf u}\|_2^2\\
{\bf z} &\leftarrow \text{arg}  ~\text{min}_{\bf z} \quad g({\bf z
}) + \rho/2 \|{\bf A}{\bf x} + {\bf B}{\bf z} - {\bf c} + {\bf u}\|_2^2\\
{\bf u} &\leftarrow {\bf u} + ({\bf A}{\bf x} + {\bf B}{\bf z} - {\bf c})
\end{aligned}
\end{equation}
\vspace{-3mm}

\noindent where ${\bf u}$ is a scaled version of the dual variable, and $\rho>0$ is a Lagrangian parameter. A comprehensive review of ADMM can be found in \cite{boyd2011distributed}.

%\vspace{-2mm}
\section{Proposed Framework: \method}
In this section, we present our proposed framework. We start with the mathematical formulation of \method, then we discuss the theoretical uniqueness of the proposed model. Next, we work out some design considerations of \method, then we derive the algorithmic solution. 

\vspace{-1mm}
\subsection{\method: Formulation}
\label{sec:formulation}
In many applications, categories of items 
exhibit a hierarchical tree structure -- as we showed in the introduction. 
For ease of notation, let us denote the embedding matrix of items resulting from NMF in \eqref{eq:NMF} as ${\bf B}_1 \in \R^{M_1 \times R}$, where $M_1$ is the number of items.   
Assume that the embeddings of the $M_1$ items (rows of ${\bf B}_1$) are the leaf nodes at the very bottom layer in a tree. A subset of items that belong to the same category is grouped together via one parent node, where the parent node is the embedding of the corresponding category. 
Assuming that the embeddings are fully inherited (replicated verbatim) from one's parent category, we can further decompose ${\bf B}_1$ into    
\begin{equation}\label{eq:tree1}
\begin{aligned}
{\bf B}_1 = {\bf S}_1{\bf B}_2
\end{aligned}
\end{equation}
where each row of ${\bf B}_2 \in \R^{M_2 \times R}$ is the embedding of one category, $M_2$ is the number of categories with $M_2 \leq M_1$, and ${\bf S}_1 \in \{0,1\}^{M_1 \times M_2}$, $\|{{\bf S}_1}(i,:)\|_0 = 1, \forall i\in[M_1]$, i.e.,
values in ${\bf S}_1$ are binary with only one 1 per row to ensure that each item belongs to exactly one category.
Note that $M_2$ is the number of parent nodes (categories) in the second from bottom layer.
The $M_2$ categories can be grouped into coarser categories, i.e., we decompose ${\bf B}_2$ into ${\bf B}_2 = {\bf S}_2{\bf B}_3$, where rows of ${\bf B}_3 \in \R^{M_2 \times M_3}$ represent the embeddings of the coarse categories, and ${\bf S}_2$ maps the $M_2$ fine-level categories into the $M_3$ coarse-level categories in the same fashion as ${\bf S}_1$. Up to here, we have constructed a three-layer tree, and we can use the same concept to create a $Q$-layer tree.  
Fig.~\ref{fig:dff} illustrates the mapping between ${\bf B}_1$ and ${\bf B}_2$ in matrix notation (left), and shows a 3-layer tree (right).
Generalizing to $Q$ layers, we obtain
\begin{equation}\label{eq:fulltree}
\begin{aligned}
{\bf B}_1 = {\bf S}_1{\bf S}_2 \dots {\bf S}_{Q-1}{\bf B}_Q
\end{aligned}
\end{equation}
Substituting the embedding matrix of items in \eqref{eq:NMF} with the right term in \eqref{eq:fulltree} above
may seem natural, however there is a solution ambiguity in the cases where $Q>2$. To see this, the route ${{\bf B}_1}{(1,:)} \rightarrow{{{\bf B}_2}{(2,:)}}$ $\rightarrow$ ${\bf B}_3(1,:)$ in Fig. \ref{fig:dff} (right) would give the same cost value as ${{\bf B}_1}{(1,:)} \rightarrow{{{\bf B}_2}{(1,:)}}$ $\rightarrow$ ${{\bf B}_3}{(1,:)}$ (the dotted gray arrow). 
Moreover, imposing the tree structure as a hard constraint can be too intrusive when the data do not exactly follow the assumed prior.    
Thus, we propose to: 
i) incorporate the tree prior as a soft constraint, and 
ii) explicitly solve for the embedding of the intermediate layers to resolve the (immaterial for our purposes) solution ambiguity.
This yields the following formulation:   
\begin{figure}[t]
	\centering
	{\includegraphics[width=0.425\textwidth,height=2.75cm]{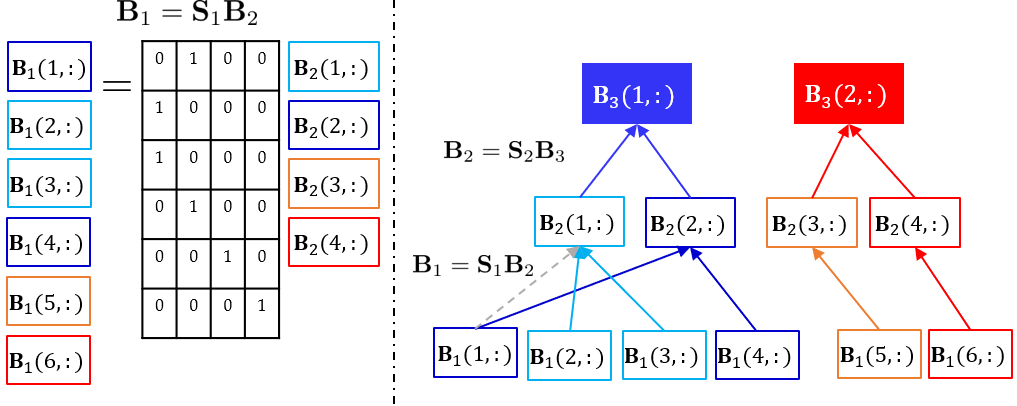}}
	\setlength{\belowcaptionskip}{-15pt}
	\vspace{-3mm}
	\caption{Illustration of the tree prior in \method.}
	\label{fig:dff}
\end{figure}
\begin{equation}\label{eq:Treehard}
\begin{aligned}
\text{min}_{\begin{subarray}{c}\mathcal{U}\end{subarray}}
\quad &  
\mathcal{F}({\bf A}, {\bf B}_1)
+ \frac{\mu}{2}  \textstyle\sum_{q=1}^{Q-1}\|{\bf B}_q - {\bf S}_q{\bf B}_{q+1}\|_F^2\\
\text{s.t.} \quad 
& {{\bf S}_q} \in \{0,1\}^{M_q \times M_{q+1}}, q \in [Q-1] \\
& \|{\bf S}_q(i,:)\|_0 = 1,  ~\forall i\in [M_q], q \in [Q-1]\\
& {\bf A} \geq {\bf 0},~{\bf B}_1 \geq {\bf 0}   
\end{aligned}
\end{equation}
where $\mathcal{U}:= \big\{{\bf A}, \{{\bf B}_q\}_{q=1}^Q, \{{\bf S}_q\}_{q=1}^{Q-1}\big\}$ is the set of all variables, and $\mathcal{F}$ is the NMF cost function as defined in \eqref{eq:NMF} with ${\bf B}_1$ as the embedding matrix of items. 
The second term is to minimize the difference between the embedding of each child node and its parent node in the tree structure. In other words, it minimizes the difference between the embeddings of each item and its category, or between 
each fine category and its coarse category. $\mu\geq 0$ is a regularization parameter to balance the data fidelity and the tree prior. 

There is an intriguing connection between the proposed tree regularizer and k-means formulation. The variables $\{{\bf S}_{q}\}_{q=1}^{Q-1}$ are equivalent to the assignment variables in k-means for clustering the rows of $\{{\bf B}_q\}_{q=1}^{Q-1}$, respectively, and ${\bf B}_Q$ is equivalent to the centroid variable in k-means for clustering the rows of ${\bf B}_{Q-1}$. On the other hand, each variable in $\{{\bf B}_q\}_{q=2}^{Q-1}$ is involved in two terms: 1) $\|{\bf B}_{q-1} - {\bf S}_{q-1}{\bf B}_{q}\|_F^2$ where its rows are centroids, and 2) $\|{\bf B}_q - {\bf S}_q{\bf B}_{q+1}\|_F^2$ where its rows are the points to be clustered. This can be thought of as a \emph{regularized} k-means. Interestingly, the joint NMF and latent k-means model in \cite{yang2016learning} is a special case of \method with $Q=2$.

Note that if the tree structure is known apriori, it can be seamlessly incorporated using our formulation. A direct way is to fix $\{{\bf S}_q\}_{q=1}^{Q-1}$ to the known tree and solve \eqref{eq:Treehard} w.r.t. the rest of the variables. This way we learn the embedding of the categories and penalize the distance between items and their corresponding categories.
When the tree is \emph{partially} known, one can fix the known parts and learn the unknowns.
Another way to integrate a known tree is to penalize the difference between the embeddings of items that share similar paths to the root nodes.    
The latter method does not require us to learn $\{{\bf B}_q\}_{q=2}^{Q}$. 
In this work, we focus on the more challenging scenario where the tree structure is unknown and to be learned from the data.

\method has the following advantages: i) it incorporates the tree structure to improve the quality of embeddings, ii) unlike most methods in the literature, it assumes the tree is unknown and learns it through the solution of $\{{\bf S}_q\}_{q=1}^{Q-1}$ in an unsupervised fashion, 
and iii) it provides the embedding of the parent nodes (categories) in addition to the item embeddings.
The tree clustering can be useful in broader applications such as classification and data labeling.  
The embeddings of categories provide extra information for some applications, e.g., web personalization  and category-based recommendation \cite{he2017category}.    

\subsection{\method: Theoretical Identifiability \label{sec:Identifiability}} 
Identifiability in machine learning problems that require parameter estimation is essential in guaranteeing sensible results,
especially in applications where model identifiability is entangled with interpretability, such as topic modeling \cite{arora2013practical}, image processing \cite{lee1999learning}, and social network clustering \cite{mao2017mixed}.   
Nevertheless, the majority of MF-based methods in practice do not have known identifiability guarantees. 
In the following theorem, we establish the identifiability of \method for the case where ${\bf X}$ is fully observed. 

\vspace{-3.5mm}
\begin{theorem}\label{the:etreeid}
Assume that a data matrix follows ${\bf X} = {\bf A}{\bf B}_{1}^T$, where ${\bf A} \in \R^{N \times R}$, and ${\bf B}_1 \in \R^{M_1 \times R}$ are the ground-truth factors, and assume that ${\bf B}_1 = {\bf S}_1 {\bf S}_2 \dots {\bf S}_{Q-1}{\bf B}_Q$, where ${\bf S}_q \in \{0,1\}^{M_q \times M_{q+1}},~\|{\bf S}_q(i,:)\|_0 = 1, \forall i \in [M_q],q \in [Q-1]$. 
Let ${\bf S} = {\bf S}_1{\bf S}_2 \dots {\bf S}_{Q-1}$, then, ${\bf B}_1 = {\bf S}{\bf B}_Q$.
Also, assume that rank$({\bf X})$ = rank$({\bf A})$ = $R$, and, without loss of generality, ${\bf M}_Q \geq R$.  
If ${\bf A}$ and ${\bf S}$ are full-column rank, 
and rows of ${\bf B}_Q$ are sufficiently scattered, then rows of ${\bf B}_1$ are sufficiently scattered, and ${\bf A}$, ${\bf B}_1$, ${\bf B}_Q$, and ${\bf S}$ are essentially unique. 
\end{theorem}
\vspace{-2mm}

\noindent {\bf Proof Sketch:} 
{The factors ${\bf A}$ and ${\bf B}_Q$ in ${\bf M} = {\bf A}{\bf B}_Q^T$ are essentially unique by Theorem \ref{theo:NMF}, since ${\bf A}$ is full-column rank, and rows of ${\bf B}_Q$ are sufficiently scattered (Definition \ref{def:ss}).
}
If ${\bf S}$ is full-column rank, then all the rows of ${\bf B}_Q$ will appear in ${\bf B}_1 = {\bf S}{\bf B}_Q$. 
Thus, rows of ${\bf B}_1$ are sufficiently scattered \emph{iff} the rows of ${\bf B}_Q$ are sufficiently scattered.
Now, ${\bf A}$ and ${\bf B}_1$ in ${\bf X} = {\bf A}{\bf B}_1^T$ are essentially unique by Theorem \ref{theo:NMF}, since ${\bf A}$ is full-column rank, and rows of ${\bf B}_1$ are sufficiently scattered.
{
Next, the factor ${\bf S}$ in ${\bf B}_1 = {\bf S} {\bf B}_Q$ is also essentially unique 
because the rows of ${\bf B}_1$ are rows of ${\bf B}_Q$, and every row of ${\bf B}_Q$ appears in ${\bf B}_1$, hence ${\bf S}$ can be determined based on the correspondence (identifiability of ${\bf S}$ also follows as a very special instance of Theorem 1). 
}

In plain words, in addition to the NMF identifiability conditions (${\bf A}$ to be full-rank, and rows of ${\bf B}_Q$ to be sufficiently scattered), we only require ${\bf S}$ to be full-column rank, which means that every root node (main category) must have at least one leaf node --- this is a natural condition in a tree.     
Interestingly, the rows of ${\bf B}_Q$ are likely to be sufficiently scattered as they are the embeddings of the coarsest categories and encouraged to be distant (think, e.g., in a 3-layer tree, each row is the centroid of multiple subcategories, where each subcategory is the centroid of a set of items). We point out that there is an inherent column permutation ambiguity in $\{{\bf S}_q\}_{q=1}^{Q-1}$ in ${\bf S} = {\bf S}_1{\bf S}_2 \dots {\bf S}_{Q-1}$, however, this is immaterial in our context.         

\subsection{\method: Model Engineering} 
\label{sec:design}
In this section, we discuss some caveats that need to be addressed in the formulation \eqref{eq:Treehard} before moving to the algorithmic derivation.

The first point is the scaling between the low-rank factors ${\bf A}$ and ${\bf B}_1$. The tree structure regularizer implicitly favors ${\bf B}_1$ to have a small norm. On the other hand, the first term is not affected by the scaling of ${\bf B}_1$, as long as this scaling is compensated for in ${\bf A}$. This motivates introducing norm regularization on ${\bf A}$, i.e., $\|{\bf A}\|_F$.

The second consideration is regarding the tree structure term.
It has been shown that the cosine similarity metric is superior over the Euclidean distance in clustering \cite{strehl2000impact} and latent clustering \cite{yang2016learning} in many applications. 
We also observed that constraining the rows of 
$\{{\bf B}\}_{q=1}^{Q-1}$ to be in the unit $l_2$-norm ball, which is equivalent to using cosine similarity in clustering, gives better performance. 
Taking these points into account, we obtain the following formulation:
\begin{equation}\label{eq:TreeD}
\begin{aligned}
\text{min}_{\begin{subarray}{c} \mathcal{Y}\end{subarray}}  
& ~~\mathcal{F}_d({\bf A}, {\bf B}, {\bf D})
+ \frac{\mu}{2} %\textstyle
\sum_{q=1}^{Q-1}\|{\bf B}_q - {\bf S}_q{\bf B}_{q+1}\|_F^2
+ \frac{\lambda}{2}\|{\bf A}\|_F^2 \\
\text{s.t.}   
& ~~ \|{{\bf B}_q}{(i,:)}\|_2 = 1 ~\forall i\in [M_q], ~\forall q\in [Q-1]\\
\quad 
& ~~{{\bf S}_q} \in \{0,1\}^{M_q \times M_{q+1}}, \forall q \in [Q-1]\\
& ~~\|{\bf S}_q(i,:)\|_0 = 1,  ~\forall i\in [M_q], q \in [Q-1]\\
& ~~{\bf D}  = \text{Diag}(d_1, \dots, d_{M_1})\\
& ~~{\bf A} \geq {\bf 0}, ~{\bf B}_1 \geq {\bf 0}  
\end{aligned}
\end{equation}
where $\mathcal{Y} := \big\{{\bf A}, {\bf D}, \{{\bf B}_q\}_{q=1}^Q, \{{\bf S}_q\}_{q=1}^{Q-1}\big\}$ is the set of all variables, 
$\lambda \geq 0$,
$\mathcal{F}_d := 1/2\|{\bf W}\odot({\bf X} -  {\bf AB}^T{\bf D})\|_F^2$, 
and ${\bf D}$ is a diagonal matrix that is introduced to allow us to fix the rows of ${\bf B}_1$ onto the unit $l_2$-norm ball without loss of generality of the factorization model.  
\subsection{\method: Algorithm}
\label{sec:algorithm}
The optimization problem in \eqref{eq:TreeD} is
NP-hard (as it contains both NMF and k-means as special cases, and both are known to be NP-hard). We therefore present a carefully designed alternating optimization (AO) algorithm.
The proposed algorithm leverages ADMM (reviewed in the background section above) and utilizes parallel computing, computation caching, and warm-start to provide a scalable implementation. 
The high level algorithmic strategy is to employ AO to update ${\bf A}$, ${\bf D}$, $\{{\bf B}_q\}_{q=1}^Q$, and $\{{\bf S}_q\}_{q=1}^{Q-1}$ one at a time, while fixing the others. The resulting sub-problems w.r.t. a single variable can be solved optimally.    

We propose a variable-splitting strategy by introducing slack variables $\{{\bf Z}_q \in \R^{M_q \times R}\}_{q=1}^{Q-1}$ to handle the unit $l_2$ norm ball constraints on $\{{\bf B}_q \in \R^{M_q \times R}\}_{q=1}^{Q-1}$ in \eqref{eq:TreeD}.   
Specifically, we consider the following optimization surrogate:
\begin{equation}\label{eq:TreeDAlg}
\begin{aligned}
\text{min}_{\begin{subarray}{c} \mathcal{H}\end{subarray}}  
\quad & %\frac{1}{2}
\mathcal{F}_d({\bf A}, {\bf B}, {\bf D})
+ \frac{\mu}{2} \textstyle\sum_{q=1}^{Q-1} \|{\bf B}_q - {\bf S}_q{\bf B}_{q+1}\|_F^2\\ 
\quad & 
+ \frac{\eta}{2} \textstyle\sum_{q=1}^{Q-1} \|{\bf B}_q - {\bf Z}_q\|_F^2
+ \frac{\lambda}{2} \|{\bf A}\|_F^2\\
\text{s.t.} 
\quad &  \|{{\bf Z}_q}{(i,:)}\|_2 = 1, \forall i \in[M_q], q\in [Q-1] \\
& {{\bf S}_q} \in \{0,1\}^{M_q \times M_{q+1}}, \forall q \in [Q-1]\\
& \|{\bf S}_q(i,:)\|_0 = 1,  ~\forall i\in [M_q], q \in [Q-1]\\
& {\bf D}  = \text{Diag}(d_1, \dots, d_{M_1})\\
& {\bf A} \geq {\bf 0}, ~{\bf B}_1 \geq {\bf 0}  
\end{aligned}
\end{equation}
where $\mathcal{H} := \big\{ {\bf A}, {\bf D}, \{{\bf B}_q\}_{q=1}^Q, \{{\bf Z}_q\}_{q=1}^{Q-1}, \{{\bf S}_q\}_{q=1}^{Q-1}\big\}$ is the set of all the variables, and $\eta \geq 0$. 
Note that when $\eta = + \infty$, then \eqref{eq:TreeDAlg} is equivalent to \eqref{eq:TreeD}. In practice, we choose a large $\eta$ to enforce ${\bf B}_q \approx {\bf Z}_q$ (we set it to $\eta = 1000$ in all experiments). 
We handle problem \eqref{eq:TreeDAlg} as follows. First, we update ${\bf A}$ by solving the following non-negative least squares %problem
\begin{equation}\label{eq:TreeDhard_A}
\begin{aligned}
\min_{\begin{subarray}{c}{\bf A} \geq {\bf 0}\end{subarray}}  \quad 
& \frac{1}{2}\|{\bf W} \odot ({\bf X} -  {\bf AB}_1^T{\bf D})\|_F^2 
+ \frac{\lambda}{2} \|{\bf A}\|_F^2\\
\end{aligned}
\end{equation}
using ADMM. 
Due to space limitation, we use the update of ${\bf A}$ as a working example for the two updates that uses ADMM (${\bf A}$ and ${\bf B}_1$). Problem \eqref{eq:TreeDhard_A} can be reformulated by introducing an auxiliary variable $\widetilde{\bf A}$
\begin{subequations}\label{eq:TreeDhard_A_admm}
\begin{align}
\min_{{\bf A}, \widetilde{\bf A}}  \quad
&\frac{1}{2}\|{\bf W}^T \odot ({\bf X}^T -  \widetilde{\bf B} \widetilde{\bf A} )\|_F^2
+ \frac{\lambda}{2} \|\widetilde{\bf A}\|_F^2 + \mathcal{R}({\bf A})\nonumber\\
\text{s.t.}  \quad &{\bf A} = \widetilde{\bf A}^T\tag{\ref{eq:TreeDhard_A_admm}}
\end{align}
\end{subequations}
%\noindent 
where $\widetilde{\bf B} = {\bf D}{\bf B}_1$, and $\mathcal{R}(.)$ is the indicator function of the nonnegative orthant.
Next, we derive the ADMM updates 
\begin{subequations}\label{admmiterA}
\begin{align}
\widetilde{\bf A}{(:,i)}  &\leftarrow ({\widetilde{\bf B}{(\mathcal{J}_i,:)}}^T\widetilde{\bf B}{(\mathcal{J}_i,:)} + c{\bf I}_R)^{-1} \big(\widetilde{\bf B}{(\mathcal{J}_i,:)}^{T} \cdot \nonumber \\ 
&\quad \quad {\bf X}{(i,\mathcal{J}_i)}^T + \rho (\widetilde{\bf A}{(:,i)} + {\bf U}{(i,:)^T})\big) \label{ADMM_1} \\ 
{\bf A}{(i,:)}  &\leftarrow [\widetilde{\bf A}{(:,i)}^T - {\bf U}{(i,:)}]_{+} \label{ADMM_2} \\
{\bf U}{(i,:)} &\leftarrow {\bf U}{(i,:)} + {\bf A}{(i,:)} - \widetilde{\bf A}{(:,i)^T} \label{ADMM_3}
\end{align}
\end{subequations}
where $c := \lambda + \rho$, $[.]_+$ is the projection on $\R_+$ by zeroing the negative entries, 
and $\mathcal{J}_i$ is the set of items that have observations for the $i^{th}$ individual. 
We use the adaptive $\rho = \|\widetilde{\bf B}\|_F^2/NR$, which is a scaled version of $\rho$ suggested in \cite{huang2016flexible}.   
The ADMM steps in \eqref{admmiterA} are performed until a termination criterion is met. We adopt the criterion in \cite{boyd2011distributed,huang2016flexible}, namely, the primal and dual residuals
\begin{equation}\label{eq:ADMMres}
\begin{aligned}
& p_i = \|{\bf A}{(i,:)} - \widetilde{\bf A}{(:,i)}^T\|_F^2 /\|{\bf A}{(i,:)}\| _F^2; \\
& d_i = \|{\bf A}{(i,:)} - {{\bf A}_0}{(i,:)}\|_F^2 / \|{\bf U}{(i,:)}\| _F^2;
\end{aligned}
\end{equation}
where ${\bf A}_0$ is ${\bf A}$ from the previous iteration. We iterate between the ADMM updates until $p$ and $d$ are smaller than a predefined threshold, or we reach the maximum number of iterations $K$ -- in our experiments we set $K=5$.

\noindent{\bf Scalability Considerations:} There are some important observations regarding the implementation of the ADMM updates in \eqref{admmiterA}. First, we do not compute the matrix inversion in \eqref{ADMM_1} explicitly. Instead, the \emph{Cholesky decomposition} of the Gram matrix ${\bf G}_i:= {\widetilde{\bf B}{(\mathcal{J}_i,:)}}^T\widetilde{\bf B}{(\mathcal{J}_i,:)} + c{\bf I}_R$ is computed, i.e., ${\bf G}_i = {\bf L}_i{\bf L}_i^T$, where ${\bf L}_i$ is a lower triangular matrix.  
Then, at each ADMM iteration, we only need to perform a forward and a backward substitution to get the solution of $\widetilde{\bf A}{(:,i)}$. Thus, the step in \eqref{ADMM_1} is replaced with:

\vspace{-3mm}
\begin{subequations}\label{admmiterAscaled}
\begin{align}
& {\bf G}_i \hspace{-1mm}\leftarrow\hspace{-1mm} {\widetilde{\bf B}}(\mathcal{J}_i,:)^T\widetilde{\bf B}{(\mathcal{J}_i,:)} \hspace{-.1mm}+\hspace{-.1mm} c{\bf I}_R; ~{\bf L}_i \hspace{-1mm}\leftarrow\hspace{-1mm} \text{Cholesky}({\bf G}_i) \label{admmiterAscaledL}\\
& {\bf F}_i \leftarrow \widetilde{\bf B}{(\mathcal{J}_i,:)}^T{\bf X}{(i,\mathcal{J}_i)}^{T}\label{admmiterAscaledF}\\
& \widetilde{\bf A}{(:,i)}  \leftarrow {\bf L}_i^{-T}{\bf L}_i^{-1}({\bf F}_i + \rho(\widetilde{\bf A}{(:,i)} + {\bf U}{(i,:)}^T)) \label{admmiterAscaledA}
\end{align}
\end{subequations}
Computing the Cholesky decomposition requires $\mathbfcal{O}(R^3)$ flops, and the back and forward substitution steps cost $\mathbfcal{O}(NR^2)$. 
The matrix multiplication in $\widetilde{\bf B}(\mathcal{J}_i,:)^T\widetilde{\bf B}{(\mathcal{J}_i,:)}$ and in computing ${\bf F}_i$ in \eqref{admmiterAscaledF} takes $\mathbfcal{O}(|\mathcal{J}_i|R^2)$ and  $\mathbfcal{O}(|\mathcal{J}_i|R)$, respectively, where, $|\mathcal{J}_i|\leq N$ is the cardinality of the set $\mathcal{J}_i$. 
An important implication is that ${\bf L}_i$ and ${\bf F}_i$ do not change throughout the ADMM iterations, thus can be cached to save computation. The overall complexity to update ${\bf A}$ is $\mathbfcal{O}(NR^2)$.  
Moreover, the ADMM updates enjoy row separability, allowing parallel computation.
In the case where ${\bf X}$ is fully observed, ${\bf G}:= {\widetilde{\bf B}^T\widetilde{\bf B}} + c{\bf I}_R$ and    
${\bf F}: = \widetilde{\bf B}^{T}{\bf X}^{T}$ are shared not only across the ADMM iterations, but also among the $N$ parallel sub-problems corresponding to the rows of ${\bf A}$. 
Finally, the outer AO routine naturally provides a good initial point (warm-start) to the inner ADMM iterations (for both ${\bf A}$ and its dual variable ${\bf U}$), resulting in a faster convergence. 

Next, we update ${\bf B}_1$ using ADMM in the same fashion as ${\bf A}$. The updates of ${\bf Z}_1$ and ${\bf D}$ admit closed form solutions
%as follows 
\begin{equation}\label{eq:ZD}
\begin{aligned}
{{\bf Z}_1}{(j,:)} %\leftarrow %\frac
= {{{\bf B}_1}{(j,:)}}/{\|{{\bf B}_1}{(j,:)}\|_2}; ~d_j %\leftarrow \frac 
= {{\bf h}_j^T{\bf X}{(\mathcal{I}_j,j)}}/{{\bf h}_j^T{\bf h}_j}
\end{aligned}
\end{equation}
where ${\bf h}_j = ({{\bf B}_1}{(j,:)}{\bf A}^T)^T$, and $\mathcal{I}_j$ is the set of individuals that have observations for the $j^{th}$ item. 

In the next step, we perform few inner iterations to alternate between updating the tree structure triplets $\{{\bf S}_{q-1}, {\bf B}_{q}, {\bf Z}_q)\}_{q=2}^{Q-1}$ and ${\bf B}_Q$ 
in a cyclic fashion (we call it the tree loop) -- in the experiments we set the maximum number of tree iterations $T=5$. 
The updates w.r.t. $\{{\bf B}_q\}_{q=2}^{Q-1}$ are unconstrained least squares problems. 
These problems are column separable with a common mixing matrix. 
Thus, the complexity can be reduced by computing \emph{one} Cholesky decomposition. Then, at each iteration, the update of each column only requires a forward and a backward substitution as follows  
%\vspace{-2mm}
\begin{subequations}\label{treeiter}
\begin{align}
 & {\bf H} \leftarrow \mu{\bf S}_{q-1}^T{\bf S}_{q-1} + v{\bf I}_{M_q}; ~{\bf L} \leftarrow \text{Cholesky}({\bf H}) \label{treeiterL}\\
 & {{\bf B}_q}(:,j)\leftarrow  {\bf L}^{-T}{\bf L}^{-1}\big(\mu{\bf S}_{q-1}{\bf B}_{q-1}(:,j) + \mu {\bf S}_{q}{\bf B}_{q+1}(:,j) \cdot \nonumber\\ 
& \quad \quad \quad + \eta {\bf Z}_q(:,j)\big) \label{treeiterBq} 
\end{align}
\end{subequations}
where $v := \mu + \eta$.  The updates of ${\bf B}_Q$ and the matrices $\{{\bf S}\}_{q=1}^{Q-1}$ are similar to solving for the centroids and the assignment variables in the k-means algorithm, respectively. Let $\mathcal{T}_{m_Q} = \{i| {{\bf S}_{Q-1}}{(i,m_Q)} = 1\}$, then each row in ${\bf B}_Q$ is 
\begin{equation}\label{admmiterBQ}
\begin{aligned}
{\bf B}_Q{(m_Q,:)}  &\leftarrow {\textstyle\sum_{i\in \mathcal{T}_{m_Q}} {\bf B}_{Q-1}{(i,:)}} /{|\mathcal{T}_{m_Q}|} 
\end{aligned}
\end{equation}
And the $i^{th}$ row of the assignment matrices is updated using
\begin{equation}\label{eq:S}
{{\bf S}_q}{(i,k)} \hspace{-1mm} \leftarrow \hspace{-1mm} 
\begin{cases} 1, & k = \text{arg} \min_{m_q} \|{{\bf B}_{q-1}}{(i,:)} - {{\bf B}_{q}}{(m_q,:)}\|_2\\
0, & \mbox{otherwise}
\end{cases}
\end{equation}

The overall algorithm is summarized in Algorithm \ref{alg:algo1}.
One nice property of the proposed algorithm is that all the updates are row separable and can be computed in a distributed fashion (with the exception of $\{{\bf B}_q\}_{q=2}^{Q-1}$, which are \emph{column} separable). 

\vspace{-2mm}
\setlength{\textfloatsep}{0pt}% Remove \textfloatsep
{\begin{algorithm}[t]%[!htb]
\DontPrintSemicolon
  %\KwInput{Your Input}
  %\KwOutput{Your output}
  {\bf Initialize}: ${\bf A}, {\bf B}_1$ $\leftarrow$ NMF;\\ $\{{\bf B}_q\}_{q=2}^Q$ and $\{{\bf S}\}_{q=1}^{Q-1}$ $\leftarrow$ random; ${\bf D}$ $\leftarrow$ ${\bf I}$; ${\bf U} \leftarrow {\bf 0}$\\ 
  %\KwData{Testing set $x$}
  \Repeat{convergence}
  {
  Compute ${\bf L}_i$, and ${\bf F}_i$, $\forall i$ using \eqref{admmiterAscaledL} and \eqref{admmiterAscaledF}\\
  Set $k = 1$ \tcp*{counter of ADMM loop}
  \While{$p_i, d_i$ in \eqref{eq:ADMMres} $> \epsilon$ and $k<K$  }  
   {	
     %\tcp*{$p_i, d_i$ are defined in \eqref{eq:ADMMres}}
        Update $\widetilde{\bf A}{(:,i)}$, ${\bf A}{(i,:)}$, and ${\bf U}{(i,:)}$, $\forall i$ using  \eqref{admmiterAscaledA}, \eqref{ADMM_2}, and \eqref{ADMM_3}, respectively\;
        $k = k + 1$\;   		
   }%\tcp*{$p$ and $d$ are defined in...}
   Update ${\bf B}_1$ using ADMM loop (similar to ${\bf A}$)\\
   Update ${\bf D}$ and ${\bf Z}_1$ using \eqref{eq:ZD}\\
   Set $t = 1$ \tcp*{counter of tree loop}
   \While{$t<T$}  
   {
           \For{$q = {2, \dots, Q-1}$}    
        { 
        	compute ${\bf L}$ using \eqref{treeiterL}\;
        	update ${{\bf B}_q}{(:,j)}$, $\forall j$ using \eqref{treeiterBq}\;
        	update ${{\bf S}_{q-1}}{(i,:)}$, $\forall i$ using \eqref{eq:S}\;
        	${{\bf Z}_q}{(i,:)} = {{\bf B}_q}{(i,:)}/\|{{\bf B}_q}{(i,:)}\|_2$, $\forall i$
        }
   update ${{\bf B}_Q}{(i,:)}$, $\forall i$ using \eqref{admmiterBQ}\;
    t = t + 1
   }
   }
\caption{Algorithmic Solution to \method}
\label{alg:algo1}
\end{algorithm}}

\begin{table*}[!htb]
	\caption{Matrix Completion Errors. {\bf \method significantly improves the prediction accuracy}.}
	\vspace{-3mm}
	\centering
	\label{table:results}
	\resizebox{0.75\textwidth}{!}{
		\begin{tabular}{c| c c| c c| c c| c c| c c| c c }
			\hline
		  & \multicolumn{2}{c}{\bf BMF} & \multicolumn{2}{|c}{\bf AdaError} & \multicolumn{2}{|c}{\bf HSR} & \multicolumn{2}{|c}{\bf NMF} &  \multicolumn{2}{|c}{\bf \method} & \multicolumn{2}{|c}{\bf NMF+KM}\\
			\hline
			\multirow{1}{*}{{\bf Data}}   & {\bf RMSE}   & {\bf MAE} & {\bf RMSE}   & {\bf MAE} & {\bf RMSE}   & {\bf MAE} &  {\bf RMSE}   & {\bf MAE} & {\bf RMSE}   & {\bf MAE} & {\bf RMSE}   & {\bf MAE}\\
			\hline 
		\multirow{1}{*}{{\bf Med-HF}}   & 0.9875  & 0.7721 & 0.9147 & 0.6858 & 0.9287 & 0.7094 & \underline{0.9031} & {\bf 0.6788} & {\bf 0.8873} & \underline{0.6808} & 1.0797 & 0.8094\\
			\hline
	\multirow{1}{*}{{\bf Med-MCI}}    & 0.8034 & 0.6232 & 0.7468 & 0.5680 & 0.7807 & 0.5990  & \underline{0.7445} & \underline{0.5612} &  {\bf 0.7317} & {\bf 0.5611} & 0.8781 & 0.6578\\
			\hline
	\multirow{1}{*}{{\bf MovieLens}}	 & 0.9300 & 0.7312  &  \underline{0.9123} & \underline{0.7165}  & 0.9216 & 0.7226 & 0.9286 &  0.7286  & {\bf 0.9106} & {\bf 0.7136} & 1.0182 & 0.8250\\
			\hline
	\multirow{1}{*}{{\bf College Grades}}	    &  0.5765 & 0.4254  & 0.5777 &  \underline{0.4206} & 0.5844 & 0.4339 & \underline{0.5755} & 0.4229 &  {\bf 0.5601} & {\bf 0.4126} & 0.5991 & 0.4476\\
	\hline
	\end{tabular}} 
	\vspace{-4mm}
\end{table*}

%\vspace{-4mm}
\section{Experiments}
In this section, we evaluate the proposed framework on real data from various application domains: healthcare analytics, movie recommendations, and education. This section aims to answer the following questions:

\noindent {\bf Q1.} {\bf Accuracy:} Does \method improve the quality of embeddings for the downstream tasks?

\noindent {\bf Q2.} {\bf Interpretability:} How meaningful is the tree structure learned by \method from an application domain knowledge viewpoint? 

%\vspace{-1mm}
%\subsubsection
\noindent{\bf Datasets:} 
We evaluate \method and the competing baselines on the following real datasets:
\noindent {(i) {\bf Med-HF:}} 
These data are provided by IQVIA Inc. and include the counts of medical services %(prescriptions, diagnosis, and procedures) 
performed on patients with heart failure (HF) conditions, including patients with preserved ejection fraction (pEF), and reduced ejection fraction (rEF).  
We include the $5,000$ patients with the most records in our experiments. 
The total number of medical services is $411$. 
The majorities of the counts fall in the range of small numbers, with a small percentage of larger numbers, resulting in a ``long tail" in the histogram of the data. To circumvent this, we apply a logarithmic transform on ${\bf X} + 1$ (we add the $1$ to be slightly above the zero as we have nonnegativity constraints). The resulting range is $log(2) - 7.79$, and the sparsity of the data matrix is $78.01\%$, 
\noindent {(ii) {\bf Med-MCI:}} These data are also provided by  IQVIA Inc. 
%This dataset is 
and similar to Med-HF, but they include patients with mild cognitive impairment (MCI) conditions. 
Similarly, we include $5,000$ patients and the total number of medical services is $412$. 
We also apply a logarithmic transform on the data. The final range of data is $log(2) - 6.98$, with a $77.76\%$ sparsity,
\noindent {(iii) {\bf Movielens:}} Movielens \cite{harper2015movielens} is a movie rating dataset %collected by GroupLens Research Project and 
and a popular baseline in recommender systems literature.
It contains $\sim 10^5$ ratings. 
The data only include users with at least $20$ ratings. We also filter out movies with less than $20$ ratings. 
The total number of users is $943$ and the total number of movies is $1,152$. 
The rating range is $1-5$, with $0.5$ increments. The sparsity of this dataset is $90.98\%$, and     
\noindent {(iv) {\bf College Grades:}} These data contain the grades of students from 
the College of Science and Engineering at the University of Minnesota 
%a specific school at a university
spanning from Fall 2002 to Spring 2013. The total number of students is $5,703$, and the number of courses is $837$. The grades take 11 discrete values ($0$, and $1$ to $4$ with increments of $1.\overline{33}$), and the sparsity of the data matrix is $96.28\%$.

%\vspace{-1mm}
%\subsubsection
\noindent{\bf Baselines:} 
We compare to the plain NMF and following state-of-the-art methods from the literature:
\noindent (i) {\bf NMF:} nonnegative Matrix factorization \eqref{eq:NMF} regularized with ($\|{\bf A}\|_F^2+\|{\bf B}\|_F^2$), and implemented using ADMM \cite{huang2016flexible}, 
{(ii) {\bf BMF}:} matrix factorization with rank-1 factors specified to capture items' and individuals' biases \cite{paterek2007improving,koren2008factorization}; implemented using Stochastic Gradient Descent (SGD). The is a well-known approach in recommender systems and considered a state-of-the-art method in student grade prediction \cite{almutairi2017context}, 
\noindent {(iii) {\bf AdaError:}} a collaborative filtering model based on matrix factorization with learning rate that adaptively adjusts based on the prediction error \cite{li2018adaerror}. AdaError is reported to have a state-of-the-art results on MovieLens \cite{rendle2019difficulty},
\noindent {(iv) {\bf HSR:}} a hierarchical structure recommender system model that captures the tree structure in items (users) via factorizing the item (user) embeddings matrix into a product of matrices, i.e., ${\bf X} = {\bf A}_1{\bf A}_2 \dots {\bf A}_{P}({\bf B}_1{\bf B}_2\dots{\bf B}_{Q})^T$ \cite{wang2018exploring, wang2015exploring}. ${\bf B}_Q \in \R^{M_{Q} \times R}$ can be interpreted as the embedding of the coarsest items categories, whereas the matrices $\{{\bf B}_q \in \R^{M_q \times M_{q+1}}\}_{q=1}^{Q-1}$ indicate the affiliation of the $M_q$ subcategories (or items) with the $M_{q+1}$ coarser categories. Note that an item can belong to all the subcategories with different scales since no constraints are imposed on ${\bf B}_q$'s matrices (except for nonnegativity). The same analysis also applies to user embeddings. 
We are unaware of other algorithms that incorporate the tree structure while learning the embeddings simultaneously.   
We used the Matlab code sample provided by the authors for a 3-layer tree and generalized it to handle Q layers, and 
{{(v) {\bf NMF+KM:}} is a simple two-stage procedure where we first apply NMF, then we obtain the embeddings of the root nodes ${\bf B}_Q$ and the product of the assignment matrices ${\bf S} = {\bf S}_1\dots {\bf S}_{Q-1}$ via k-means' centroids and assignment variable, respectively --  we include NMF+KM to demonstrate the advantage  of learning the embeddings and tree structure simultaneously.      
}
 
%\vspace{-2mm}
%\subsubsection
\noindent{\bf Q1. Accuracy of Embeddings:}
The quality of embeddings can be evaluated by testing their performance with a particular task, e.g., classification or regression. Here we take a more generic approach and evaluate the embedding quality on matrix completion. The philosophy is: \emph{if the embeddings predict missing data with high accuracy, then they must be good representations of items and individuals.}  
We split each dataset into 5 equal folds. 
After training the models on 4 folds ($80\%$ of the data), we test the trained models on the held-out fold.  
The hyper-parameters of all methods are chosen via cross validation ($10\%$ of training data). 
Due to random initialization, the results can differ for different runs. Thus, after choosing the hyper-parameters, we run the training and testing on each fold $20$ times and report the average error of the total $100$ experiments. 
       
Table \ref{table:results} shows the Root Mean Square Error (RMSE) and Mean Absolute Error (MAE) of all methods on the different datasets. We highlight the smallest error in bold and underline the second smallest. 
\method significantly improves the best baseline with all datasets. Note that MovieLens and the grade datasets are challenging and an improvement in the second digit is considered significant in the literature.       
By comparing NMF and \method, we can conclude that the tree prior enhances the accuracy.
Moreover, we can see the clear advantage of simultaneously learning the embeddings and tree clusters when we compare \method with MNF+KM.
We point out that HSR baseline works better with MovieLens, compared to the medical datasets. This is likely because a movie usually belongs to a mix of genres, which suits the complete tree assumption in HSR. Nevertheless, the proposed tree formulation in \method provides better accuracy.        

\begin{figure}[t]%[htbp]%[t]
	\centering
	%\vspace{-1mm}
	\subfloat[Sample of the 412 medical services in Med-MCI.]
	{\includegraphics[width=0.475\textwidth,height=3cm]{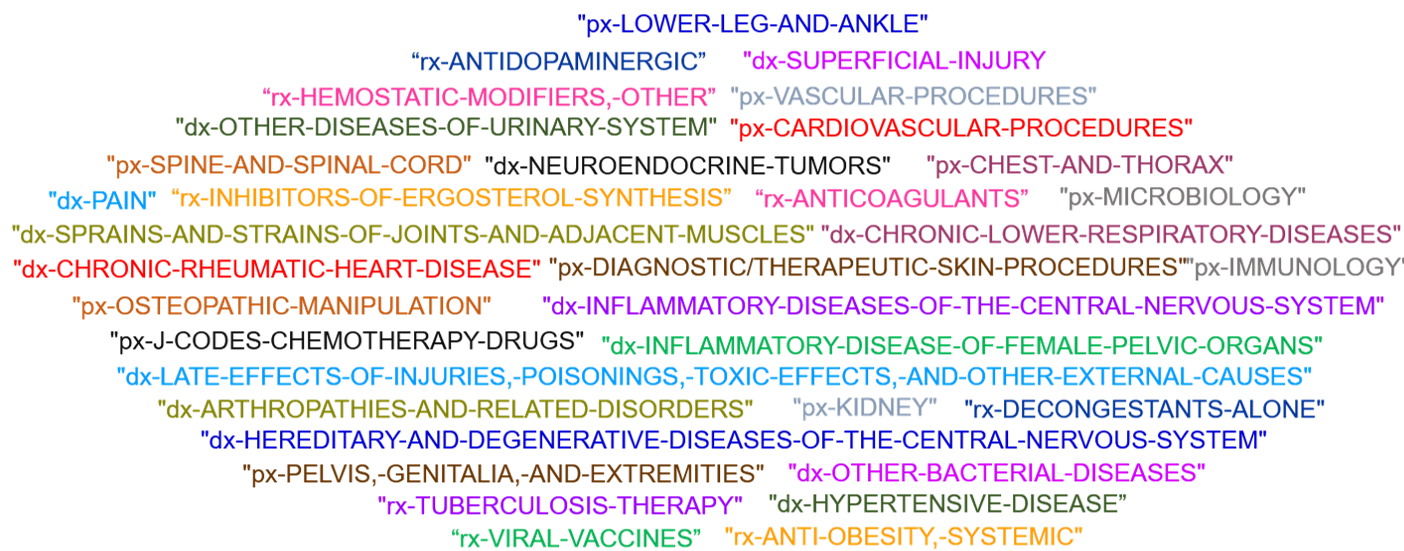}}\label{fig:R1a}
	\subfloat[A tree learned by \method and labeled by domain experts.]
	{\includegraphics[width=0.475\textwidth,height=3.5cm]{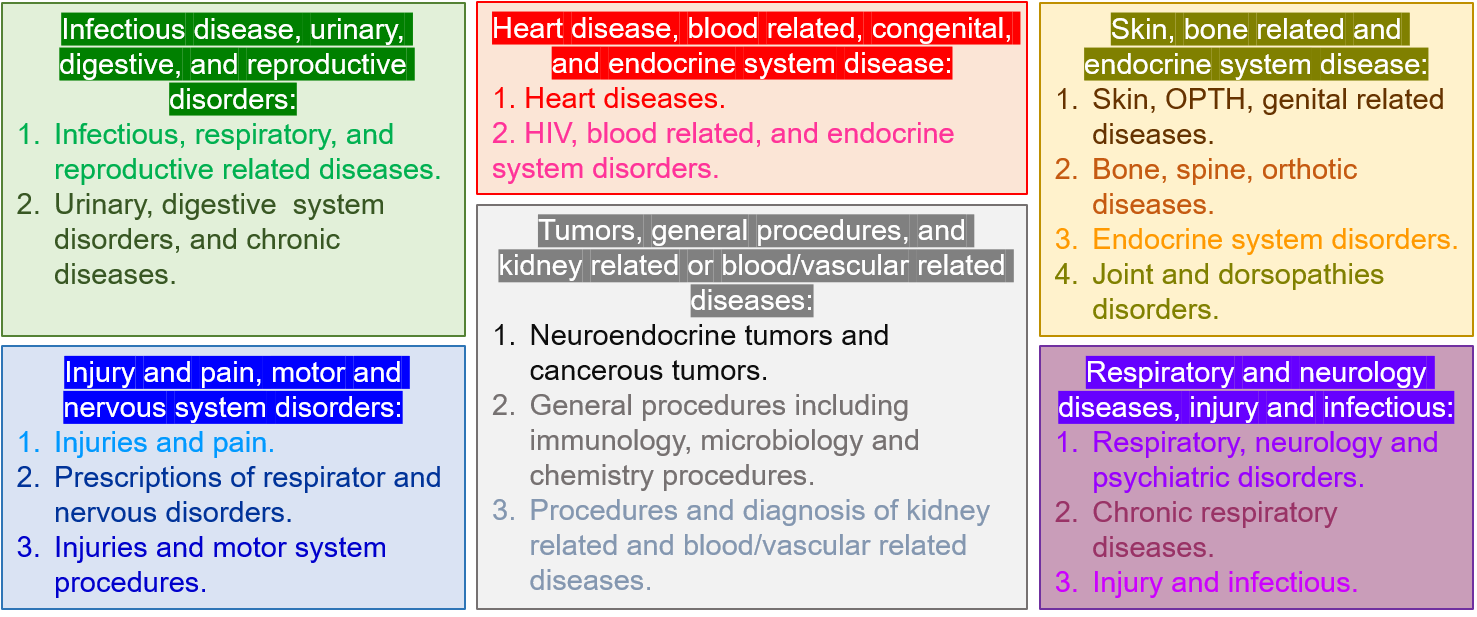}}\label{fig:R1b}
		\vspace{-3mm}
	\caption{\bf \method Provides Meaningful Clusters.}
	\label{fig:Med_MCI}
\end{figure}

%\vspace{-2mm}
%\subsubsection
\noindent{\bf Q2. Interpretability of Learned Trees:}
{We ran a 3-layer \method on Med-MCI with the following parameters: $R=9$, $\lambda=1$, $\mu=50$ (more emphasis on the tree term), $M_2 = 27$ (number of subcategories), and $M_3 = 9$ (number of main categories). A sample of the $412$ medical services is shown in Fig. \ref{fig:Med_MCI}~(a), {where `dx' stands for a diagnosis, `rx' is a prescription, and `px' is a procedure.}
{
Note that \method assigns each medical service to one subcategory, and each subcategory to a main category. 
Services with the \emph{same} color in Fig. \ref{fig:Med_MCI}~(a) belong to the same subcategories, whereas services with \emph{similar} colors (e.g., light and dark blue) belong to the same {main} category but to different subcategories. 
}
These unsupervised tree clusters 
%of the medical services learned by \method 
were {then} shown to medical professionals. 
The domain experts were able to find coherence in the groups and {they labeled} both the main categories and their subcategories. 
The tables in Fig. \ref{fig:Med_MCI}~(b) shows the names of the main categories and their subcategories {as labeled by the medical professionals} -- we show the top 6 coherent categories. 
Similar interpretability was observed on Med-HF data, but not shown due to space limitation.}

\vspace{-1mm}
\section{Conclusion}
In this paper, we proposed \method, a framework that incorporates the tree structure while learning the embeddings of a data matrix. \method not only exploits the tree structure, but also learns the hierarchical clustering in an supervised fashion. 
We leveraged the special properties of NMF to prove the uniqueness of the proposed model. We employed ADMM, parallel computing, and computation caching to derive a lightweight algorithm with scalable implementation to solve \method. 
We showed the effectiveness of \method on real data from healthcare, recommender systems, and education domains.  
The interpretability of \method was demonstrated using real data and proved by domain experts.

%\textcolor{red}{done till here}

%In the first experiment, we aim to evaluate the tree structure learning of the proposed model. We generate synthetic data according to:
%\begin{equation}\label{eq:TreeDhard_A}
%\begin{aligned}
%{\bf X}  =   {\bf A}{\bf B}_1 + {\bf E}
%\end{aligned}
%\end{equation}
%where ${\bf B}_1 = {\bf S}_1{\bf B}_2 + {\bf E}_1$, and $\{{\bf B}_q\}_{q=2}^{Q-1}$ are defined similarly. The matrices ${\bf E}$, and $\{{\bf E}_q\}_{q=1}^{Q-1}$ denote the measurement noise and modeling errors, respectively. We generate a factor ${\bf A}$ from i.i.d. standard Gaussian distribution, with all negative entries set to zero. Then, the factor ${\bf B}_{Q}$, that carries the embedding of tree top nodes, is generated by setting the first $R$ rows as unit vectors (to ensure identifiability of the NMF model), i.e., ${\bf B}_Q(1:R,:) = {\bf I}$, and entries in the remaining $M_Q - R$ are randomly generated from an i.i.d. uniform distribution in $[0,1]$. 

%\begin{table}[htbp]%[!htb]
%	\caption{Tree (clustering) Accuracy for the Case where $Q=2$}
%	\centering
%	\label{table:scalability}
	%\resizebox{0.5\textwidth}{!}{
%		\begin{tabular}{c|| c |c |c |c |c}
%			\hline
%		SNR[dB]   & {3} & {6} & { 9} & { 12} & noiseless\\
%			\hline
%			\hline
%			{\bf NTL} (proposed)   & 81.4  &  82.6 & 85.8 & 90 & 100\\
%			\hline
%			{\bf KM}    &   48.2 & 62.2 & 69.6 & 82.6 & 93\\
%	\end{tabular}%} 
%\end{table}

\bibliography{ref.bib}

\begin{thebibliography}{33}
\providecommand{\natexlab}[1]{#1}
\providecommand{\url}[1]{\texttt{#1}}
\providecommand{\urlprefix}{URL }
\expandafter\ifx\csname urlstyle\endcsname\relax
  \providecommand{\doi}[1]{doi:\discretionary{}{}{}#1}\else
  \providecommand{\doi}{doi:\discretionary{}{}{}\begingroup
  \urlstyle{rm}\Url}\fi

\bibitem[{Almutairi, Sidiropoulos, and Karypis(2017)}]{almutairi2017context}
Almutairi, F.~M.; Sidiropoulos, N.~D.; and Karypis, G. 2017.
\newblock Context-aware recommendation-based learning analytics using tensor
  and coupled matrix factorization.
\newblock \emph{IEEE Journal of Selected Topics in Signal Processing} 11(5):
  729--741.

\bibitem[{Arora et~al.(2013)Arora, Ge, Halpern, Mimno, Moitra, Sontag, Wu, and
  Zhu}]{arora2013practical}
Arora, S.; Ge, R.; Halpern, Y.; Mimno, D.; Moitra, A.; Sontag, D.; Wu, Y.; and
  Zhu, M. 2013.
\newblock A practical algorithm for topic modeling with provable guarantees.
\newblock In \emph{International Conference on Machine Learning}, 280--288.

\bibitem[{Boyd, Parikh, and Chu(2011)}]{boyd2011distributed}
Boyd, S.; Parikh, N.; and Chu, E. 2011.
\newblock \emph{Distributed optimization and statistical learning via the
  alternating direction method of multipliers}.
\newblock Now Publishers Inc.

\bibitem[{Fan and Cheng(2018)}]{fan2018matrix}
Fan, J.; and Cheng, J. 2018.
\newblock Matrix completion by deep matrix factorization.
\newblock \emph{Neural Networks} 98: 34--41.

\bibitem[{Fu, Huang, and Sidiropoulos(2018)}]{fu2018identifiability}
Fu, X.; Huang, K.; and Sidiropoulos, N.~D. 2018.
\newblock On identifiability of nonnegative matrix factorization.
\newblock \emph{IEEE Signal Processing Letters} 25(3): 328--332.

\bibitem[{Fu et~al.(2015)Fu, Ma, Huang, and Sidiropoulos}]{fu2015blind}
Fu, X.; Ma, W.-K.; Huang, K.; and Sidiropoulos, N.~D. 2015.
\newblock Blind separation of quasi-stationary sources: Exploiting convex
  geometry in covariance domain.
\newblock \emph{IEEE Transactions on Signal Processing} 63(9): 2306--2320.

\bibitem[{Harper and Konstan(2015)}]{harper2015movielens}
Harper, F.~M.; and Konstan, J.~A. 2015.
\newblock The movielens datasets: History and context.
\newblock \emph{Acm transactions on interactive intelligent systems (tiis)}
  5(4): 1--19.

\bibitem[{He, Li, and Liao(2017)}]{he2017category}
He, J.; Li, X.; and Liao, L. 2017.
\newblock Category-aware Next Point-of-Interest Recommendation via Listwise
  Bayesian Personalized Ranking.
\newblock In \emph{IJCAI}, volume~17, 1837--1843.

\bibitem[{Hoyer(2004)}]{hoyer2004non}
Hoyer, P.~O. 2004.
\newblock Non-negative matrix factorization with sparseness constraints.
\newblock \emph{Journal of machine learning research} 5(Nov): 1457--1469.

\bibitem[{Huang, Sidiropoulos, and Liavas(2016)}]{huang2016flexible}
Huang, K.; Sidiropoulos, N.~D.; and Liavas, A.~P. 2016.
\newblock A flexible and efficient algorithmic framework for constrained matrix
  and tensor factorization.
\newblock \emph{IEEE Transactions on Signal Processing} 64(19): 5052--5065.

\bibitem[{Koren(2008)}]{koren2008factorization}
Koren, Y. 2008.
\newblock Factorization meets the neighborhood: a multifaceted collaborative
  filtering model.
\newblock In \emph{Proceedings of the 14th ACM SIGKDD international conference
  on Knowledge discovery and data mining}, 426--434.

\bibitem[{Lee and Seung(1999)}]{lee1999learning}
Lee, D.~D.; and Seung, H.~S. 1999.
\newblock Learning the parts of objects by non-negative matrix factorization.
\newblock \emph{Nature} 401(6755): 788--791.

\bibitem[{Li et~al.(2018)Li, Chen, Lv, Gu, Lu, Shang, Gu, and
  Chu}]{li2018adaerror}
Li, D.; Chen, C.; Lv, Q.; Gu, H.; Lu, T.; Shang, L.; Gu, N.; and Chu, S.~M.
  2018.
\newblock Adaerror: An adaptive learning rate method for matrix
  approximation-based collaborative filtering.
\newblock In \emph{Proceedings of the 2018 World Wide Web Conference},
  741--751.

\bibitem[{Li et~al.(2019)Li, Liu, Qian, Mamoulis, Tu, and Cheung}]{li2019hhmf}
Li, H.; Liu, Y.; Qian, Y.; Mamoulis, N.; Tu, W.; and Cheung, D.~W. 2019.
\newblock HHMF: hidden hierarchical matrix factorization for recommender
  systems.
\newblock \emph{Data Mining and Knowledge Discovery} 33(6): 1548--1582.

\bibitem[{Lin et~al.(2015)Lin, Ma, Li, Chi, and
  Ambikapathi}]{lin2015identifiability}
Lin, C.-H.; Ma, W.-K.; Li, W.-C.; Chi, C.-Y.; and Ambikapathi, A. 2015.
\newblock Identifiability of the simplex volume minimization criterion for
  blind hyperspectral unmixing: The no-pure-pixel case.
\newblock \emph{IEEE Transactions on Geoscience and Remote Sensing} 53(10):
  5530--5546.

\bibitem[{Liu et~al.(2015)Liu, Luo, Tao, Xu, and Wen}]{liu2015low}
Liu, M.; Luo, Y.; Tao, D.; Xu, C.; and Wen, Y. 2015.
\newblock Low-rank multi-view learning in matrix completion for multi-label
  image classification.
\newblock In \emph{Proceedings of the Twenty-Ninth AAAI Conference on
  Artificial Intelligence}, 2778--2784.

\bibitem[{Maleszka, Mianowska, and Nguyen(2013)}]{maleszka2013method}
Maleszka, M.; Mianowska, B.; and Nguyen, N.~T. 2013.
\newblock A method for collaborative recommendation using knowledge integration
  tools and hierarchical structure of user profiles.
\newblock \emph{Knowledge-Based Systems} 47: 1--13.

\bibitem[{Mao, Sarkar, and Chakrabarti(2017)}]{mao2017mixed}
Mao, X.; Sarkar, P.; and Chakrabarti, D. 2017.
\newblock On mixed memberships and symmetric nonnegative matrix factorizations.
\newblock In \emph{International Conference on Machine Learning}, 2324--2333.

\bibitem[{Menon et~al.(2011)Menon, Chitrapura, Garg, Agarwal, and
  Kota}]{menon2011response}
Menon, A.~K.; Chitrapura, K.-P.; Garg, S.; Agarwal, D.; and Kota, N. 2011.
\newblock Response prediction using collaborative filtering with hierarchies
  and side-information.
\newblock In \emph{Proceedings of the 17th ACM SIGKDD international conference
  on Knowledge discovery and data mining}, 141--149.

\bibitem[{Nikolakopoulos, Kouneli, and
  Garofalakis(2015)}]{nikolakopoulos2015hierarchical}
Nikolakopoulos, A.~N.; Kouneli, M.~A.; and Garofalakis, J.~D. 2015.
\newblock Hierarchical itemspace rank: Exploiting hierarchy to alleviate
  sparsity in ranking-based recommendation.
\newblock \emph{Neurocomputing} 163: 126--136.

\bibitem[{Paterek(2007)}]{paterek2007improving}
Paterek, A. 2007.
\newblock Improving regularized singular value decomposition for collaborative
  filtering.
\newblock In \emph{Proceedings of KDD cup and workshop}, volume 2007, 5--8.

\bibitem[{Rendle, Zhang, and Koren(2019)}]{rendle2019difficulty}
Rendle, S.; Zhang, L.; and Koren, Y. 2019.
\newblock On the difficulty of evaluating baselines: A study on recommender
  systems.
\newblock \emph{arXiv preprint arXiv:1905.01395} .

\bibitem[{Shen et~al.(2018)Shen, Tan, Sordoni, and Courville}]{shen2018ordered}
Shen, Y.; Tan, S.; Sordoni, A.; and Courville, A. 2018.
\newblock Ordered Neurons: Integrating Tree Structures into Recurrent Neural
  Networks.
\newblock In \emph{International Conference on Learning Representations}.

\bibitem[{Strehl, Ghosh, and Mooney(2000)}]{strehl2000impact}
Strehl, A.; Ghosh, J.; and Mooney, R. 2000.
\newblock Impact of similarity measures on web-page clustering.
\newblock In \emph{Workshop on artificial intelligence for web search (AAAI
  2000)}, volume~58, 64.

\bibitem[{Sun et~al.(2017)Sun, Yang, Zhang, and Bozzon}]{sun2017exploiting}
Sun, Z.; Yang, J.; Zhang, J.; and Bozzon, A. 2017.
\newblock Exploiting both vertical and horizontal dimensions of feature
  hierarchy for effective recommendation.
\newblock In \emph{Thirty-First AAAI Conference on Artificial Intelligence}.

\bibitem[{Trigeorgis et~al.(2016)Trigeorgis, Bousmalis, Zafeiriou, and
  Schuller}]{trigeorgis2016deep}
Trigeorgis, G.; Bousmalis, K.; Zafeiriou, S.; and Schuller, B.~W. 2016.
\newblock A deep matrix factorization method for learning attribute
  representations.
\newblock \emph{IEEE transactions on pattern analysis and machine intelligence}
  39(3): 417--429.

\bibitem[{Wang et~al.(2015)Wang, Tang, Wang, and Liu}]{wang2015exploring}
Wang, S.; Tang, J.; Wang, Y.; and Liu, H. 2015.
\newblock Exploring Implicit Hierarchical Structures for Recommender Systems.
\newblock In \emph{IJCAI}, 1813--1819.

\bibitem[{Wang et~al.(2018)Wang, Tang, Wang, and Liu}]{wang2018exploring}
Wang, S.; Tang, J.; Wang, Y.; and Liu, H. 2018.
\newblock Exploring hierarchical structures for recommender systems.
\newblock \emph{IEEE Transactions on Knowledge and Data Engineering} 30(6):
  1022--1035.

\bibitem[{Wang, Pan, and Xu(2014)}]{wang2014hgmf}
Wang, X.; Pan, W.; and Xu, C. 2014.
\newblock Hgmf: Hierarchical group matrix factorization for collaborative
  recommendation.
\newblock In \emph{Proceedings of the 23rd ACM International Conference on
  Conference on Information and Knowledge Management}, 769--778.

\bibitem[{Wang et~al.(2019)Wang, Wu, Wang, and Wang}]{wang2019enhancing}
Wang, Y.; Wu, T.; Wang, Y.; and Wang, G. 2019.
\newblock Enhancing model interpretability and accuracy for disease progression
  prediction via phenotype-based patient similarity learning.
\newblock In \emph{Pac Symp Biocomput}. World Scientific.

\bibitem[{Yang, Fu, and Sidiropoulos(2016)}]{yang2016learning}
Yang, B.; Fu, X.; and Sidiropoulos, N.~D. 2016.
\newblock Learning from hidden traits: Joint factor analysis and latent
  clustering.
\newblock \emph{IEEE Transactions on Signal Processing} 65(1): 256--269.

\bibitem[{Yang et~al.(2016)Yang, Sun, Bozzon, and Zhang}]{yang2016learningHier}
Yang, J.; Sun, Z.; Bozzon, A.; and Zhang, J. 2016.
\newblock Learning hierarchical feature influence for recommendation by
  recursive regularization.
\newblock In \emph{Proceedings of the 10th ACM Conference on Recommender
  Systems}, 51--58.

\bibitem[{Zhong, Fan, and Yang(2012)}]{zhong2012contextual}
Zhong, E.; Fan, W.; and Yang, Q. 2012.
\newblock Contextual collaborative filtering via hierarchical matrix
  factorization.
\newblock In \emph{Proceedings of the 2012 SIAM International Conference on
  Data Mining}, 744--755. SIAM.

\end{thebibliography}

\end{document}